\documentclass[10pt,letterpaper]{article}
\usepackage[top=0.85in,left=1.25in,footskip=0.75in,marginparwidth=2in]{geometry}

\usepackage[utf8]{inputenc}

\usepackage{cite}

\usepackage{nameref,hyperref}

\usepackage[right]{lineno}

\usepackage{microtype}
\DisableLigatures[f]{encoding = *, family = * }

\raggedright
\usepackage{changepage}

\usepackage[aboveskip=1pt,labelfont=bf,labelsep=period,singlelinecheck=off]{caption}

\makeatletter
\renewcommand{\@biblabel}[1]{\quad#1.}
\makeatother

\usepackage{lastpage,fancyhdr,graphicx}
\usepackage{epstopdf}
\fancyheadoffset[L]{2.25in}
\fancyfootoffset[L]{2.25in}

\usepackage{color}

\definecolor{Gray}{gray}{.25}

\usepackage{graphicx}

\usepackage{sidecap}

\usepackage{wrapfig}
\usepackage[pscoord]{eso-pic}
\usepackage[fulladjust]{marginnote}
\reversemarginpar

\begin{document}
\vspace*{0.35in}

\begin{flushleft}
{\Large
\textbf\newline{A dataset of 40K naturalistic 6-degree-of-freedom robotic grasp demonstrations.}
}
\newline
\\
Rajan Iyengar,
Victor Reyes Osorio,
Presish Bhattachan,
Adrian Ragobar,
Bryan Tripp\textsuperscript{*}
\\
\bigskip
\bf University of Waterloo
\\
\bigskip
* bptripp@uwaterloo.ca

\end{flushleft}

\section*{Abstract}
Modern approaches to grasp planning often involve deep learning. However, there are only a few large datasets of labelled grasping examples on physical robots, and available datasets involve relatively simple planar grasps with two-fingered grippers. Here we present: 1) a new human grasp demonstration method that facilitates rapid collection of naturalistic grasp examples, with full six-degree-of-freedom gripper positioning; and 2) a dataset of roughly forty thousand successful grasps on 109 different rigid objects with the RightHand Robotics three-fingered ReFlex gripper. 


\section*{Introduction}
Grasp planning has traditionally used analytic methods to estimate quality metrics for potential grasps \cite{Ferrari1992,Miller1999}. Two particular limitations of analytic grasp metrics are the need for accurate knowledge of the object geometry, and assumptions involved, such as simplified contact models. Much recent work in grasp planning has focused on data-driven approaches \cite{Bohg2014} to address both of these limitations. A common approach is to use deep learning to map depth or RGB images to quantities that can be used directly for grasp planning, such as grasp success predictions \cite{Saxena2008,Lenz2015,Kappler2015,Wang2016,Pinto2016,Mahler2017,Schmidt2018}. For example, uncertainty about the shapes of novel objects has been addressed by training deep networks to predict grasp metrics from depth images, using large numbers of known synthetic examples \cite{Mahler2017}. However, the relationship between these metrics and physical grasp success is complex \cite{Rubert2017}. 

Ideally, a grasp planner would be trained directly on physical grasping examples rather than grasp quality metrics. However, a central challenge for this approach is the expense of obtaining sufficient labelled data to support sophisticated decisions without overfitting. To reduce this expense, some groups have resorted to large-scale physics simulations \cite{kleinhans2015g3db,Kappler2015}. However, these simulations have (so far) employed simplified contact models, reintroducing one of the key limitations that motivated a departure from grasp quality metrics. Other groups have trained models initially in simulated environments, then trained further with physical robots \cite{Bousmalis2018,zhu2018reinforcement}, or performed large-scale trial-and-error data collection \cite{Pinto2016} or reinforcement learning \cite{Gu2017,Levine2018,Quillen2018} on physical robots. Datasets of successful grasps have also been generated by humans. This requires either transfer from human-hand grasps to robotic-gripper grasps \cite{ekvall2007learning,lin2015robot} or human control of a robot. The latter has been achieved by physically guiding the robotic arm \cite{akgun2012keyframe}, and recently, teleoperation using virtual reality hardware \cite{zhang2018deep}. These methods have, so far, not produced sufficiently large datasets for deep learning of skilled grasping (although this seems feasible with teleoperation). Overall, while robotic grasp planning with unknown objects has been extensively studied, there is still much room for improvement in success rates. New labelled datasets of physical robotic grasps may allow further progress.

We developed a new grasp demonstration approach that is intended to make grasp demonstration relatively rapid and naturalistic. Here we describe this approach, and a corresponding dataset of 40K successful grasps, demonstrated on 109 objects. We believe this to be the largest available dataset of human grasp demonstrations with a robotic gripper. The grasps use a three-fingered gripper (RightHand Robotics' ReFlex gripper), and full 6-degree-of-freedom trajectories. 

The dataset can be used, shared, and modified freely for any non-commercial purpose. It is available from https://dataverse.scholarsportal.info/dataverse/uw-brain-lab. 

\section*{Methods}
\subsection*{Grasp Demonstration Method}
\begin{figure}
    \centering
    \includegraphics[width=.6\textwidth]{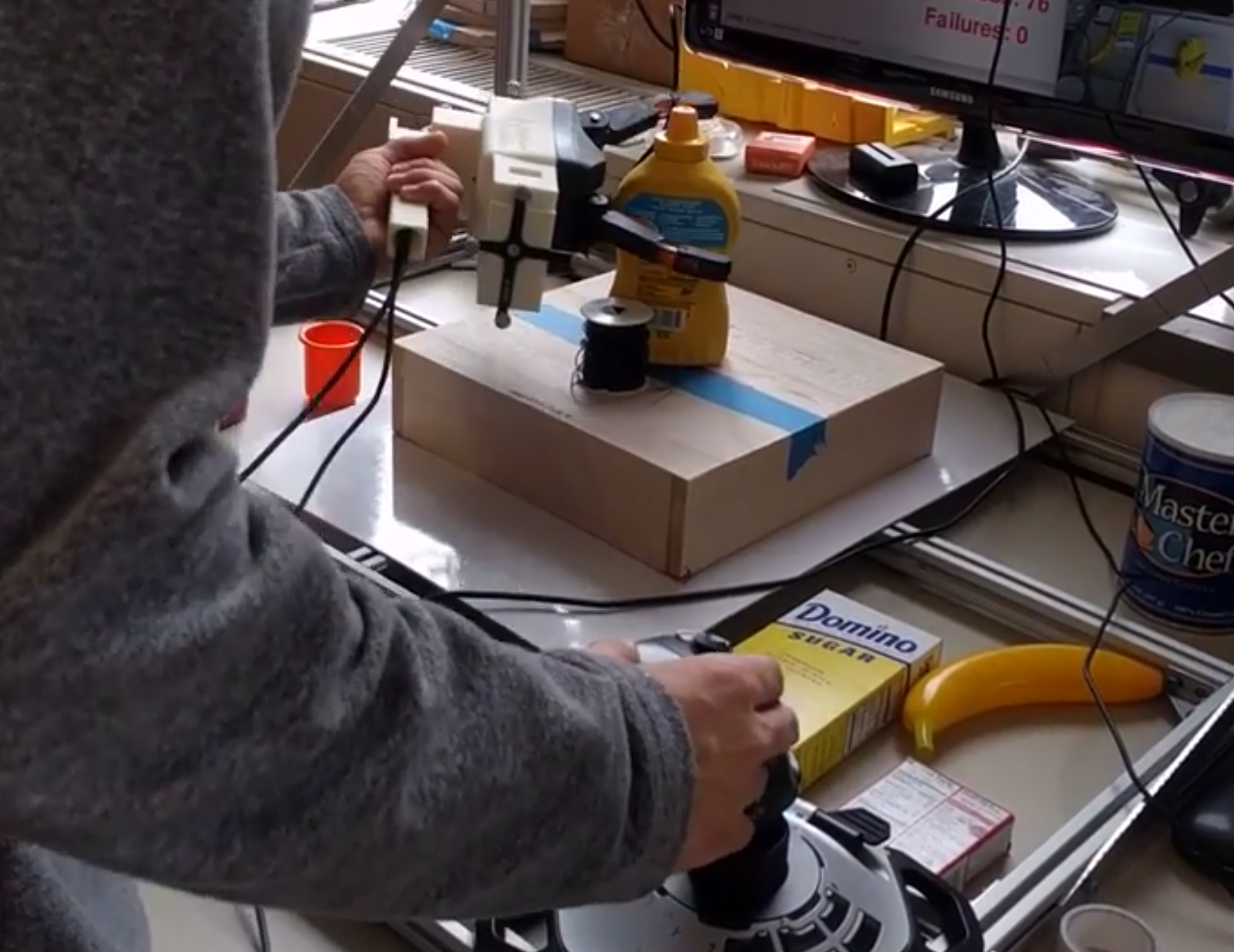}
    \vspace{1em}
    \centering
    \includegraphics[width=.4\textwidth]{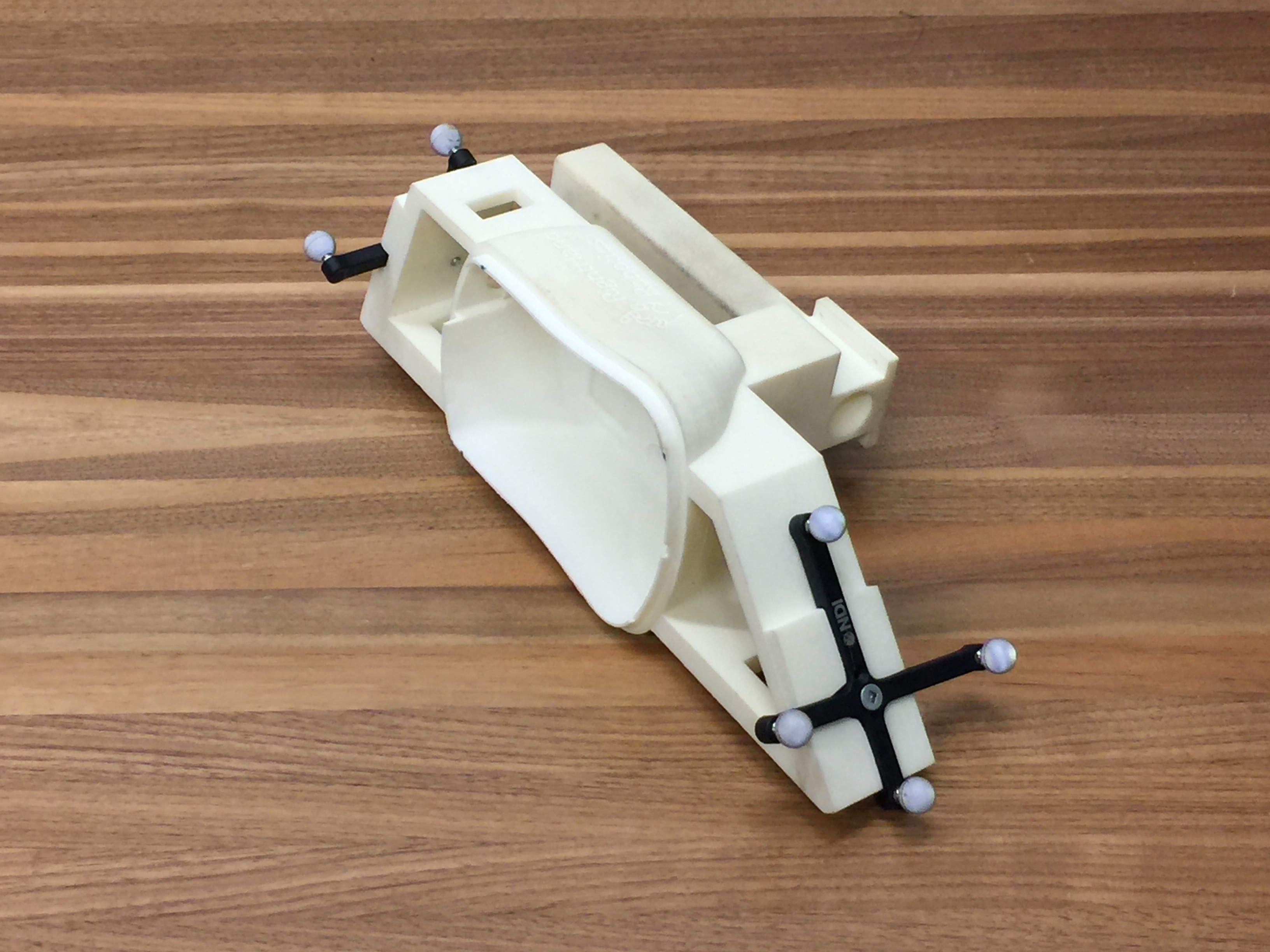}
    \includegraphics[width=.225\textwidth]{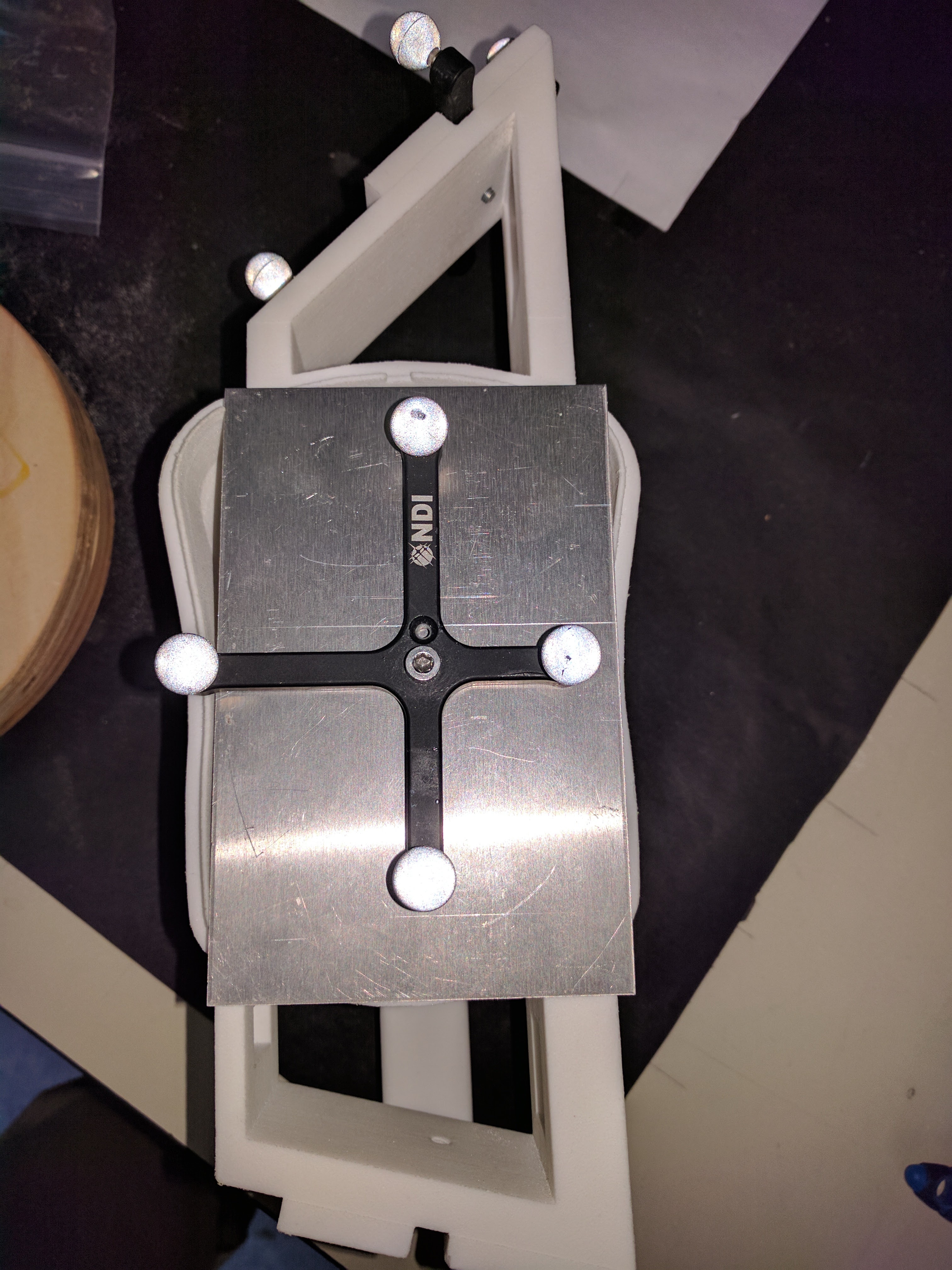}
    \caption{Top: An experimenter using the handle-mounted gripper and joystick to grasp an object. Left: 3D printed handle for naturalistic human positioning of the ReFlex gripper. This part replaces the ReFlex gripper's case. Two distinct NDI tools are mounted at known positions and orientations relative to the gripper axes, allowing us to reliably determine the gripper's configuration during grasping. Right: To measure the tool positions relative to the gripper reference frame, we temporarily mounted a tool at the point we defined as the gripper origin, and aligned with its axes.}
    \label{fig:demonstration-hardware}
\end{figure}

Our method is a hybrid of previous approaches that have used motion tracking with human-hand grasps \cite{ekvall2007learning}, and manual control of a robot \cite{akgun2012keyframe}. To approach the speed and naturalistic control of human-hand grasping while avoiding the need to generalize from the hand to a gripper, we mounted a gripper on a 3d-printed handle with motion-tracking markers (Figure \ref{fig:demonstration-hardware}). This allowed the operator to position the gripper with natural arm movements. We used a Polaris optical motion tracker from Northern Digital Inc. (NDI). This system can track the 6DOF configuration of unique multi-marker ``tools'', but it requires a line of sight to all of the markers on a tool. To reduce occlusions, we mounted two of these tools at different positions and angles on the gripper handle (Figure \ref{fig:demonstration-hardware}, bottom left). We used a RightHand Robotics ReFlex gripper. The operator controlled the gripper fingers with a joystick. This gripper has four degrees of freedom in the finger positions, corresponding to flexion of each finger, and the angle of spread between two of the fingers. We used one degree of freedom of the joystick to control the spread, and the other to control all finger flexion angles together. 

\subsection*{Objects and Data Collection}
We collected grasp demonstrations with 109 different rigid objects. 48 of these belonged to the YCB dataset \cite{Calli2015}. We avoided YCBs object that were either non-rigid, or too small or too large to easily handle with the ReFlex gripper.   

Two experimenters participated in each data collection session. In each trial, one experimenter placed an object on a table, in a random orientation, and the other used the joystick and handle to grasp and lift the object (see Figure \ref{fig:demonstration-hardware}). To encourage more variability in the grasps, one or two additional objects (``obstacles'') were placed between the operator and the target object in some trials. Failure to grasp and lift an object was rare, particularly after the operator had encountered a given object a few times. Failed grasps were not included in the dataset. 

Prior to each grasp, we captured images of the target object with two cameras. One was an infrared structured-light sensor that captured RGB and depth images (RealSense SR300). Because the motion tracker also used infrared light, this camera and the motion tracker were enabled at alternating times. The second camera was a stereo camera (Stereolabs ZED). From this camera we saved stereo RGB images, as well as the depth map estimated from these images by the ZED software. Both cameras were fixed to a rigid frame that also held a small table on which the objects were placed.
For each grasp, we stored the 6DOF trajectory of the gripper as it approached the object, along with the gripper finger positions. The gripper and finger positions were recorded on different computers, with different sampling rates. To create an integrated 10-dimensional gripper-configuration signal, we synchronized the clocks and interpolated the finger positions at the times of the motion-tracker samples.

The coordinate systems of the gripper and the table were right-handed. The positive $z$ axis of the gripper pointed out of the palm, and the positive $y$ axis pointed between the two fingers on one side of the gripper. Finger positions were recorded in units of 1/4096 rotations of the corresponding servos. A home finger position was also saved for each trial. In the home position, the two fingers on one side of the gripper were oriented parallel to each other, and the fingers were extended, so that the fingers on each side were oriented approximately 180 degrees from each other about the $x$ axis. The fingers were visually aligned to this position periodically, as well as immediately after occasional mechanical problems with the gripper that affected the relationship between servo and finger positions (such as after replacing stripped gears). 


\subsection*{Re-creating and Replaying Grasps}
After data collection, we re-created the target-object placement for some of the grasps. To do this, we overlaid the object image captured during the trial with a live image from the camera, and manually aligned the images by moving the object on the table. In some cases, we attached a motion-tracker tool to the object and manually measured its position relative to the object centre. This allowed us to analyze gripper positions in object coordinates rather than table coordinates for these grasps. 
In other cases, we replayed recorded grasps, with the gripper mounted on a robotic arm (Universal Robots UR5). This allowed us to confirm that re-creation of the object positions was fairly accurate, and to study robustness of the demonstrated grasps by replaying them with small perturbations.

\begin{figure}
    \centering
    \includegraphics[width=.5\textwidth]{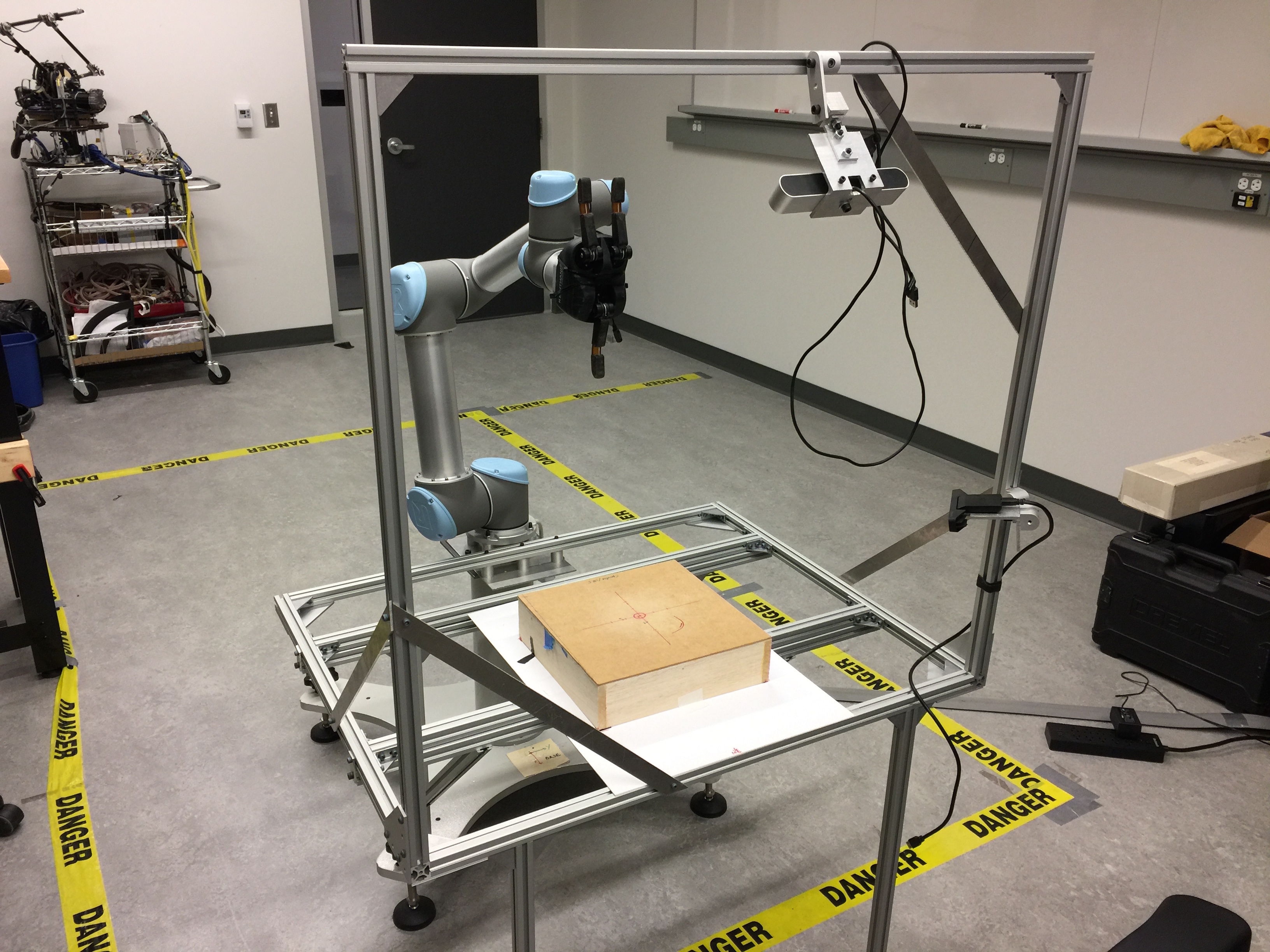}
    \caption{Setup for grasp replay on the UR5 robot. The frame is the same one used for data collection. For replay the frame is bolted to the UR5's base, whereas for data collection it is clamped to a table in front of the NDI Polaris.}
    \label{fig:replay-setup}
\end{figure}

\section*{Results}
Table \ref{tab:example-record} shows an example of a full processed record for a single grasping trial. Most of the record consists of lines with 11 numbers each. Each of these lines corresponds to a time point in the trial. The first is the time, in seconds. The next six describe the position and orientation of the gripper base, and the last four describe the finger positions (see details in Methods). The units are 1/4096-rotation increments of the servos. The first two values are flexion positions of the two fingers on one side of the gripper, the third is the flexion position of the opposed finger, and the fourth is the spread between fingers 1 and 2.

\begin{figure}
    \centering
    \includegraphics[width=.5\textwidth]{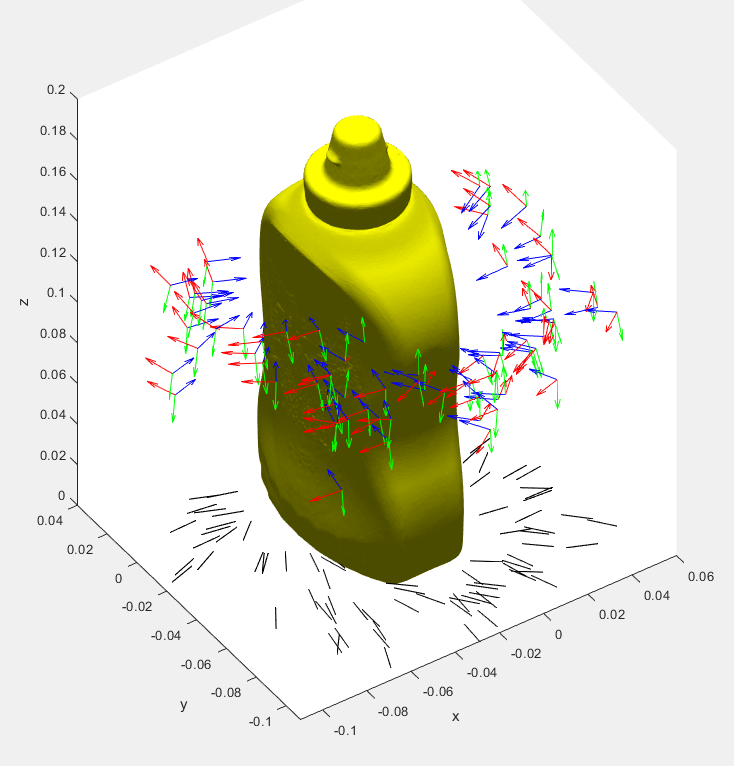}
    \caption{A render of a mustard bottle, with plots of final gripper positions and orientations (in the object's reference frame) for a number of grasps. The red, green, and blue axes are x, y, and z, respectively. The lines at the bottom are projections of the z axes onto the support surface.}
    \label{fig:mustard-examples}
\end{figure}

Figure \ref{fig:mustard-examples} illustrates a number of final gripper positions (for different trials) around a single object. 

\begin{table}[]
\begin{footnotesize}
\begin{verbatim}
58.766572,221.789299,-609.763151,195.410166,0.449433,-3.041296,-0.346504,15360,14073,17234,15977
58.852577,226.062290,-609.847780,194.122006,0.463341,-3.029768,-0.341773,15360,14073,17234,15977
58.932581,235.837668,-610.406901,190.827514,0.478962,-3.017942,-0.332029,15360,14073,17234,15977
59.016586,243.962991,-607.974445,190.105599,0.472996,-2.990546,-0.332048,15360,14073,17234,15977
59.101591,245.909832,-603.582400,192.819763,0.460992,-2.966361,-0.351199,15360,14073,17234,15977
59.182596,243.874769,-585.271146,208.594179,0.413329,-2.880503,-0.436490,15360,14073,17234,15977
59.718626,138.905502,-251.135107,335.640953,-0.294284,-1.607775,-0.849032,15360,14073,17234,15977
59.800631,128.693773,-216.504463,318.502908,-0.347747,-1.580540,-0.812127,15360,14073,17234,15977
59.882636,107.239968,-158.570112,275.121700,-0.417064,-1.567641,-0.740564,15360,14073,17234,15977
59.966641,87.909652,-117.351661,233.379995,-0.464845,-1.583969,-0.675839,15360,14073,17234,15977
60.051645,80.553602,-102.145953,218.497081,-0.481372,-1.568064,-0.655075,15360,14073,17234,15977
60.133650,70.049938,-79.689731,199.347674,-0.484544,-1.535028,-0.640559,15360,14073,17234,15977
60.216655,61.995151,-63.102111,187.352547,-0.511094,-1.511234,-0.618503,15360,14073,17234,15977
60.301660,59.427056,-56.254151,181.030151,-0.526766,-1.511101,-0.607948,15360,14073,17234,15977
60.383664,56.054238,-46.715327,170.989789,-0.539463,-1.502067,-0.604229,15360,14073,17234,15977
60.466669,53.731411,-40.008332,164.390846,-0.559969,-1.495705,-0.594769,15360,14073,17234,15977
60.550674,52.684224,-38.058058,161.073913,-0.562399,-1.508526,-0.592344,15360,14073,17234,15977
60.632679,50.528583,-36.798384,155.019788,-0.565769,-1.531361,-0.584006,15360,14073,17234,15977
60.716683,48.763014,-36.471269,153.736346,-0.580545,-1.516428,-0.576254,15360,14073,17234,15977
60.800688,48.144880,-35.951091,153.747681,-0.588292,-1.507040,-0.571679,15360,14073,17234,15977
60.882693,46.467974,-34.466787,153.331707,-0.592097,-1.498633,-0.565814,15360,14073,17234,15977
60.966698,44.244102,-32.376847,152.113245,-0.592320,-1.504597,-0.558035,15360,14073,17234,15977
61.050703,42.687376,-31.328490,151.058910,-0.589851,-1.511646,-0.552103,15360,14073,17234,15977
61.132707,39.011888,-29.567549,148.645548,-0.573705,-1.527906,-0.542928,15360,14073,17234,15977
61.216712,35.034849,-26.950631,147.070403,-0.570720,-1.528795,-0.531688,15360,14073,17234,15978
61.301717,33.156350,-25.321857,146.550399,-0.574164,-1.523476,-0.524586,15360,14073,17234,15978
61.383721,29.951727,-22.735748,144.591925,-0.581185,-1.518510,-0.511573,15360,14073,17234,15978
61.466726,28.670496,-21.602918,143.546636,-0.589383,-1.508460,-0.503629,15497,13935,17371,16020
61.551731,28.989980,-20.890682,143.836360,-0.595603,-1.501013,-0.504591,15672,13760,17547,16072
61.632736,29.602992,-20.291707,144.032218,-0.599700,-1.499684,-0.507331,15839,13593,17714,16121
61.716741,29.722951,-20.295574,143.725106,-0.604394,-1.498913,-0.504594,16012,13420,17887,16173
61.800746,29.739973,-19.953561,143.821686,-0.610408,-1.494604,-0.503381,16185,13247,18060,16224
61.882750,30.383383,-18.717478,144.782633,-0.619521,-1.482919,-0.507993,16354,13078,18229,16275
61.966755,30.977656,-17.384658,146.225541,-0.630547,-1.473200,-0.505784,16528,12904,18402,16326
62.051760,31.556563,-17.371589,150.831706,-0.629501,-1.462754,-0.513835,16613,12820,18488,16352
62.132764,33.859368,-18.337561,168.907829,-0.607408,-1.460303,-0.552683,16613,12820,18488,16352
62.216769,37.116328,-21.036063,185.732048,-0.572061,-1.471607,-0.592257,16613,12820,18488,16352
62.300774,38.880719,-23.174612,194.285943,-0.559710,-1.459379,-0.606706,16613,12820,18488,16352
62.383779,40.771197,-27.263256,208.944944,-0.540954,-1.463413,-0.634165,16613,12820,18488,16352
T:3-medium wrap plastic spiral
2 obs
Fingers at calibration = [14523, 14910, 16398, 16365]
\end{verbatim}
\end{footnotesize}
    \caption{Complete contents of an example grasp record file (147104). The final three lines are: manually entered notes on the object and grasp type; the number of obstacles; and finger servo positions at the home position (other finger positions should be interpreted relative to these values). The preceding lines each have 11 entries each. The first is a time stamp, in seconds. Entries 2 to 4 are $x$, $y$, $z$ offsets from the centre of the table surface, respectively, in mm. Entries 5-7 are an axis-angle representation of the gripper orientation. Entries 8 to 11 are positions of the gripper fingers (see text for details).}
    \label{tab:example-record}
\end{table}

\section*{Conclusion}
We extended previous grasp demonstration methods to allow support rapid collection of a large dataset of naturalistic grasp examples. Our dataset includes roughly forty thousand examples of successful human-controlled grasps with a three-fingered gripper and a wide variety of objects. We hope this dataset will be useful for training deep networks for grasp planning, and for understanding human grasping strategies.


\section*{Acknowledgments}
We thank Ricky Verma and Ibrahim Okeil for their assistance with data collection, Xueyang Yao for assistance with data processing, and ReFlex Robotics for technical assistance. This work was supported by Applied Brain Research and NSERC. 

\nolinenumbers

\bibliography{grasp-demonstration}

\bibliographystyle{abbrv}

\end{document}